\documentclass[runningheads]{llncs}
\usepackage{graphicx}
\usepackage{comment}
\usepackage{amsmath,amssymb} % define this before the line numbering.
\usepackage{color}

\usepackage[pagebackref=true,breaklinks=true,letterpaper=true,colorlinks,bookmarks=false]{hyperref}

\usepackage{mmstyle}
\usepackage{booktabs} % To thicken table lines

\usepackage{pifont}% http://ctan.org/pkg/pifont

\usepackage{subfigure}

\begin{document}
% \linenumbers
\pagestyle{headings}
\mainmatter
\def\ECCVSubNumber{1514}  % Insert your submission number here

\title{Online Multi-modal Person Search in Videos} % Replace with your title

\begin{comment}
\titlerunning{ECCV-20 submission ID \ECCVSubNumber} 
\authorrunning{ECCV-20 submission ID \ECCVSubNumber} 
\author{Anonymous ECCV submission}
\institute{Paper ID \ECCVSubNumber}
\end{comment}
%******************

%\begin{comment}
\titlerunning{Online Multi-modal Person Search in Videos}
%\author{Jiangyue Xia\inst{1}\orcidID{0000-0003-3981-5373} \and
%Anyi Rao\inst{2}\thanks{Corresponding author} \and
%Qingqiu Huang\inst{2}\orcidID{0000-0002-6467-1634} \and
%Linning Xu\inst{2} \and \\
%Jiangtao Wen\inst{1} \and
%Dahua Lin\inst{2}\orcidID{0000-0002-8865-7896}
%}
\author{Jiangyue Xia\inst{1} \and
Anyi Rao\inst{2}\thanks{Corresponding author} \and
Qingqiu Huang\inst{2} \and
Linning Xu\inst{2} \and \\
Jiangtao Wen\inst{1} \and
Dahua Lin\inst{2}
}

\authorrunning{J. Xia, A. Rao, Q. Huang, L. Xu, J. Wen, D. Lin}
\institute{Department of Computer Science and Technology, Tsinghua University \\ \email{xiajy16@mails.tsinghua.edu.cn, jtwen@tsinghua.edu.cn} \and 
CUHK-SenseTime Joint Lab, The Chinese University of Hong Kong\\
\email{\{anyirao, hq016, dhlin\}@ie.cuhk.edu.hk, linningxu@link.cuhk.edu.cn}
}
%\end{comment}
%******************
\maketitle

%%%%%%%%% BODY TEXT
% !TEX root = ../main.tex

\begin{abstract}

The task of searching certain people in videos has seen increasing potential in real-world applications, such as video organization and editing. Most existing approaches are devised to work in an offline manner, where identities can only be inferred after an entire video is examined. This working manner precludes such methods from being applied to online services or those applications that require real-time responses. In this paper, we propose an online person search framework, which can recognize people in a video on the fly. This framework maintains a multi-modal memory bank at its heart as the basis for person recognition, and updates it dynamically with a policy obtained by reinforcement learning. Our experiments on a large movie dataset show that the proposed method is effective, not only achieving remarkable improvements over online schemes but also outperforming offline methods.

\end{abstract}

% !TEX root = ../main.tex

\section{Introduction}
\label{sec:introduction}

% (1) person search is important \\
% (2) most are off-line and online is necessary \\
% (3) online is challenging \\
% (4) our basic idea \\
% (5) tech work flow and present some results \\
% (6) summary of contributions  \\

Person identification in videos can be specified into different forms and tasks. Among them, \emph{person search with one portrait} is especially related to real-world applications, such as ``intelligent fast forwards" on online video platforms and multimedia-oriented web search, and can further benefit video summarization and story understanding. This task is very challenging compared with other person identification problems such as \emph{person Re-ID}~\cite{reid_2005_ICRA,reid_2006_CVPR,reid_2010_CVPR} and \emph{person recognition in photo album}~\cite{lin2010joint,album_2015_CVPR}, as the appearance, pose, and clothing of the characters may vary dramatically through the videos.
To overcome this difficulty, the research community has explored the use of various modalities~\cite{face_2005_CVPR,Buffy_2006_BMVC,whoru_2009_CVPR,dialog_2010_CVPR,dialog_2016_WACV,Benedict_2017_BMVC,Huang_2018_ECCV}, such as face, lip motion, body, audio, subtitle, and screenplay.

However, those methods are mainly offline,~\ie an instance is compared with the rest to determine its identity, 
which leads to high computational complexity. Additionally, for scenarios such as suspect discovery in real-time surveillance videos and story understanding in live broadcasting,
the offline approaches cannot recognize the identities immediately. In this paper, we work on \emph{online} person search to meet the emerging requirement of timely inference.

\begin{figure}[!t]
	\centering
	\includegraphics[width=0.95\linewidth]{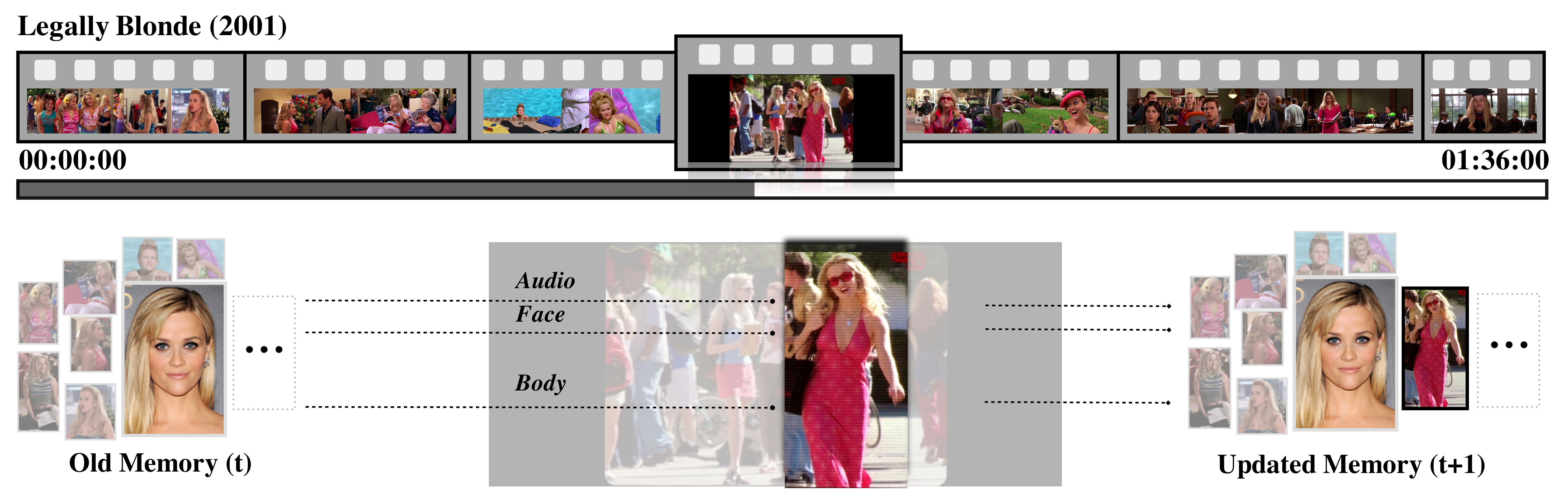}
	\caption{Illustration of the memory updating scheme of human movie watching experience. We select out instances of \emph{Elle Woods} in movie \emph{Legally Blonde} (2001) and demonstrate how we update our memory about actress \emph{Reese Witherspoon} with them. The multi-modal memory stores face, body and audio information, which are closely related to human identities}
	\label{fig:teaser}
\end{figure}

Online search is very challenging, as decisions need to be made on the fly based on limited memory. The key to this problem is to effectively update the memory so that it can adapt to the changes as the video proceeds. Think about how human tackle with online person search.
Suppose we are watching the movie \emph{Legally Blonde} (2001), as shown in Figure~\ref{fig:teaser}. 
When we see the figure of \emph{Elle Woods}, we compare it with previous images stored in our memories to infer the actress's name. There are two possibilities. 
1) If the instance appears to be very similar to \emph{Reese Witherspoon}, we recognize her name %\emph{Reese Witherspoon} 
immediately, and update the impression of \emph{Reese Witherspoon} in our memories with the current looking of \emph{Elle Woods}. Similar processes are also carried out for other cast.
When another new instance comes, we continue to compare it with our dynamically updated memory to judge his/her identity. 
2) The other possible reaction is that we cannot confirm her identity since her looking is quite different from any cast that exists in our memories. In this case, we stay confused until she appears again and again. We gradually build up our memories on \emph{Elle Woods} and may be capable of recognizing her as \emph{Reese Witherspoon} in the future.

Inspired by this cognitive process, we propose an \emph{online multi-modal searching machine} (OMS). 
Specifically, to mimic how human recognize characters and store representations in memory, a \emph{dynamic memory bank} is developed to store \emph{face}, \emph{body} and \emph{audio} features of each cast. These multi-modal feature representations are closely related to human identities.
The memory bank is dynamically updated to capture the latest changes to the cast's features as new instances come in. To adapt to diverse movie contents and appearance changes, instead of interacting with the memory by a hand-crafted rule, we formulate the process as a decision making problem and design a controller to learn the strategy of memory updating. 
Motivated by the second case we mentioned above, it is possible that an instance cannot be recognized as any cast in list at the very beginning, since the initial dynamic memory bank lacks adequate information.
We develop an \emph{uncertain instance cache} to keep these temporarily confusing instances for judgments later on. 
As the online process goes on, more and more instances are recognized and the dynamic memory bank becomes more informative, we select out instances in the cache and make a second decision for them.

Experiments are conducted on \emph{Cast Search in Movies} dataset~\cite{Huang_2018_ECCV} to verify the effectiveness of our online multi-modal searching method. Thanks to the adaptive multi-modal feature integration and reinforcement learning based memory updating strategy, our approach raises the mAP from 61.24\% to 69.08\% and outperforms all the online methods. Surprisingly, it achieves better results than offline methods and declines computational cost at the same time.

% !TEX root = ../main.tex

\section{Related Work}
\label{sec:related}

\paragraph{\bf Person Identification in Videos.}
In order to identify characters in videos, frameworks using diverse features have been proposed. What commonly used are visual features of face~\cite{face_2005_CVPR} and body~\cite{Huang_2018_ECCV,huang2020movienet}, audio features of speaking voice~\cite{Benedict_2017_BMVC}, text features of subtitle~\cite{dialog_2010_CVPR,dialog_2016_WACV} and screenplay~\cite{Buffy_2006_BMVC,whoru_2009_CVPR}, and contextual features of scene and social relation~\cite{huang2018cvpr}. In~\cite{Buffy_2006_BMVC,whoru_2009_CVPR}, with the alignment of subtitles and screenplay, time-stamped annotations are acquired to provide supervision of character naming.
Nagrani \etal.~\cite{Benedict_2017_BMVC} train face and voice classifiers in a joint framework to recognize characters. With face and body features, Huang \etal.~\cite{Huang_2018_ECCV,huang2020movienet}
propagate identity labels through visual and temporal links between the instances.
However, most previous studies work on an offline manner, \ie all the instances are compared with each other, and the corresponding identities are inferred after an entire video is examined, which increases computational complexity.
In this paper, we propose an online framework that dynamically updates the memory with features of newly identified instances to enable real-time inference. Since text information such as subtitles and screenplay is more difficult to acquire compared with the internal features, we utilize face, body and audio features to infer identities.

\paragraph{\bf Multi-modal Fusion.} 
In person identification methods, fusion of visual and audio features can be classified into two categories: late integration~\cite{classifier_2005_TMM,rao2020local} and early integration~\cite{speaker_2015_MM,temporal_2016_CVPR,LSTM_2016_AAAI,Zhou_2019_ICCV}. Late integration methods design a specific classifier for each modality and combine decisions by voting or scoring, while early integration merges features from different modalities by concatenation, weighted summation, or learning joint presentations, etc., before decision. Erzin \etal.~\cite{classifier_2005_TMM} determine the reliable modality combinations with a cascade of classifiers.
Hu \etal.~\cite{speaker_2015_MM,LSTM_2016_AAAI} propose a cross-modality weight sharing LSTM to capture correlation of face and audio features for speaker identification. In this paper, %we leverage reinforcement learning to learn the strategy of multi-modal fusion.
the strategy of multi-modal fusion is learnt implicitly in the decision making process.

\paragraph{\bf Memory Modelling.} 
To strengthen the ability of conventional neural networks in modelling long-range temporal dependencies, several memory models are proposed. Graves \etal.~\cite{ntm_2014} design a Neural Turing Machine (NTM) which holds an external memory to interact with the neural networks through attentional reading and writing operations. While NTM focuses on problems of sorting, copying and recall, Memory Networks~\cite{memnet_2015_ICLR} utilize large long-term static external memory and target to language and reasoning tasks. Sukhbaatar \etal.~\cite{memn2n_2015_NIPS} extend the model to a continuous form to enable end-to-end training, making it more generally applicable to tasks with less supervision.  These memory models have also been modified to different structures~\cite{adaptmem_2018_ICLR,ordermem_2019_NIPS} and adopted in video-related researches such as summarization~\cite{videosum_2018_MM,videosum_2019_MM}, captioning~\cite{caption_2018_MM}, visual question answering~\cite{moviestory_2017_ICCV} and object tracking~\cite{objtrack_2018_ECCV}. In this paper, we utilize a dynamic memory bank to store updated multi-modal features of cast in movies.

\paragraph{\bf Reinforcement Learning.} Reinforcement learning (RL) is a technique for solving decision making problems, aiming at learning a policy for the sequence of state-action pairs to obtain maximal rewards~\cite{RL_1998_NN}. In recent years, RL has been applied in person Re-ID~\cite{rlid_2018_MM,rlid_2019_NNLS,rlid_2019_ICCV} and face recognition~\cite{rlfr_2017_ICCV}. In~\cite{rlfr_2017_ICCV,rlid_2018_MM,rlid_2019_NNLS}, RL is used to %decide the misleading and confounding frames and 
find the most representative frames in video sequences, while in~\cite{rlid_2019_ICCV}, RL guides an agent to select informative training samples which are used to finetune a pre-trained Re-ID model. In this paper, we formulate the %fusion of multi-modal features and 
updating of memory as a decision making problem, where we learn the strategy with RL to maximize recognition accuracy.
% !TEX root = ../main.tex
\section{Online Multi-modal Search}

\begin{figure}[!t]
	\centering
	\includegraphics[width=0.85\linewidth]{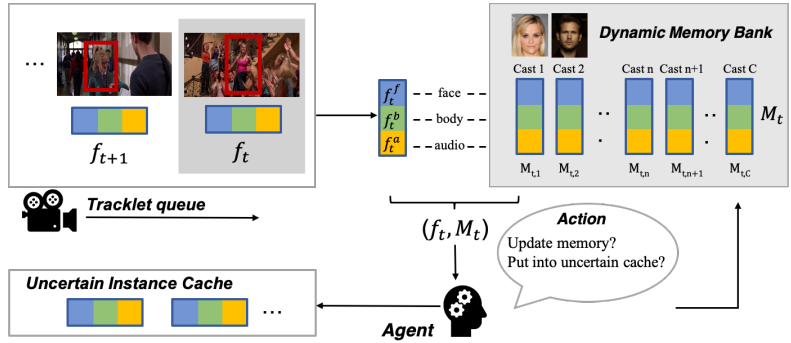}
	\caption{Pipeline of inference in our proposed OMS. A dynamic memory bank stores the multi-modal feature representations of each actor/actress. When a new instance comes, we compare it with each candidate cast, then the trained agent decides whether to update his memory with this instance or to put it into the uncertain instance cache}
	\label{fig:pipeline}
\end{figure}

Given the portraits of a list of cast, our goal is to search them in a sequential movie with an online fashion following the human behaviors.
To tackle this challenging problem, we propose a novel \textit{online multi-modal searching machine} (OMS)
as shown in Figure~\ref{fig:pipeline}.
There are four key components in OMS,
\ie multi-modal feature representations (MFR), a dynamic memory bank (DMB), an uncertain instance cache (UIC) and a controller.
Each instance is a tracklet and is represented by multi-modal features. It is compared with the cast stored in the memory bank to judge its identity.
The controller then determines whether this instance should be used to update memory or put into the uncertain instance cache for later comparisons. The memory bank and the uncertain instance cache are dynamically updated over time, with a strategy operated by the controller.
All these components together build an ``intelligent machine'' to watch a movie and gradually recognize the characters like humans do.

\subsection{Multi-modal Feature Representations}
When watching a movie, we can identify a person based on various cues, \eg~facial appearance, clothing, and even speech. These modalities are complementary to each other. 
Therefore, it is necessary for us to capture the representations of different modalities for each instance in the movie.
Specifically, we take face, body and audio information into consideration in our framework.
Given an instance $x$, we represent it with three feature vectors $(f^f(x), f^b(x), f^a(x))$.
Here $f^f \in \mathbb{R}^d$ is the face feature that comes from a face recognition model, $f^b \in \mathbb{R}^d$ is the body feature obtained by a Re-ID network,
and $f^a \in \mathbb{R}^d$ is the audio feature acquired by a speech recognition model.
These feature vectors are concatenated to form a holistic representation $f = [f^f, f^b, f^a] \in \mathbb{R}^{3d}$.

\subsection{Dynamic Memory Bank}

A simple way to search cast is to calculate the similarity between the given portrait and the detected instances by their face features.
However, as the movie proceeds, the appearance of a cast may change dramatically, and a clear face is missing in many cases where the body is partially occluded or even blurred.
Humans can tackle this problem easily with the help of memory.
Imagine that when you watch a movie, you may not be able to recognize some of the people at the beginning.
However, with the playing of the video, you become more and more familiar with the characters as more identified instances enter the memory.

Inspired by the above observation,
we construct a \emph{dynamic memory bank} (DMB) $\cM_t \in \mathbb{R}^{C \times 3d}$ to store the most representative features of each person.
Here $t \in [1, \cdots, N]$ represents the time when the $t$-th instance appears, $N$ is the total number of instances in a movie and $C$ denotes the number of cast in list.
The memory bank is initialized with the features of the provided portrait of each actor/actress.
When an instance $x_t$ comes, we search for it in our memory and then predict its identity.
The procedure can be formulated as Eq.~\ref{eq:pred},
where $f_t$ is the multi-modal feature representation of $x_t$.
\begin{equation}
	p_t = \cM_t \cdot f_t^T 
	\label{eq:pred}
\end{equation}

As the movie goes by, the DMB keeps updating, with the strategy shown as Eq.~\ref{eq:mem-update}.
Here $\mu \in [0,1]$ is a pre-defined updating factor.
$\mathcal{G}^1_{t,j} \in \{0, 1\}$ is a gate of the controller,
the details of which will be introduced in Sec.~\ref{subsec:controller},
and $j \in [1, \cdots, C]$ represents the $j$-th cast.

\begin{equation}
\cM_{t+1,j} = (1 - \mu \mathcal{G}^1_{t,j})  \cM_{t,j} + \mu \mathcal{G}^1_{t,j} f_t
\label{eq:mem-update}
\end{equation}

\subsection{Uncertain Instance Cache}

At the beginning of a movie, we are not familiar with the characters.
Therefore, it may be quite hard for us to recognize some of the tough samples.
For example, if a man appears in the first frame of the movie without a visible face,
it is impossible for us to identify him at that time.
However, as the movie goes on, we begin to know more about the story and the people.
We may suddenly recall the uncertain instance before and recognize him with our stronger knowledge.

Motivated by the fact described above,
we build a novel module in our machine to store the uncertain instances temporarily,
which is named as \emph{uncertain instance cache} (UIC).
We denote the cache as $\cC \in \mathbb{R}^{k \times 3d}$.
$k$ is the size of the cache, which dynamically changes as time goes on.
Whether to place an instance $x_t$ into the cache or not is also represented by a gate of the controller,
denoted as $\mathcal{G}_t^2  \in \{0, 1\}$, which will be introduced in Sec.~\ref{subsec:controller}.
The updating strategy can be formulated as Eq.~\ref{eq:cache-update}.
\begin{equation}
\cC_k = f_t, ~~ k \leftarrow k+1 ~~~~~~ \text{if} ~~ \mathcal{G}_t^2 = 1
\label{eq:cache-update}
\end{equation}

Whenever the DMB updates, we recall all the instances in the UIC to make new predictions.
Specifically, we compare each instance $x_i$ in the cache with the updated memory bank $\cM_t$, as shown in Eq.~\ref{eq:cache-pred}. $\cC_i$ ($i \in [1,\cdots, k]$) is the multi-modal feature representation of $x_i$. 

\begin{equation}
p_i = \cM_t \cdot \cC_i^T 
\label{eq:cache-pred}
\end{equation}

The $p_i$ here is not the final prediction of the uncertain instance $x_i$.
Whether $x_i$ can be confidently identified and removed from the cache is controlled by the third gate $\mathcal{G}_i^3 \in \{0, 1\}$,
the details of which will also be introduced in Sec.~\ref{subsec:controller}.

\subsection{Controller}
\label{subsec:controller}

As we mentioned before, there are three gates,
\ie $\mathcal{G}_{t,j}^1, \mathcal{G}_t^2, \mathcal{G}_i^3 \in \{0, 1\}$,
in our framework.
The three gates determine ``whether to update the memory with instance $x_t$'',
``whether to put $x_t$ into the uncertain cache'',
and ``whether to remove $x_i$ from the cache'', respectively. 
In this section, we will provide details on how to construct a controller with all these three gates.

\paragraph{\bf A Manual Controller.}
A simple way is to design the gates by setting thresholds for the prediction, \ie the similarity.
Eq.~\ref{eq:manual-control} shows such a manual controller,
where $\alpha$, $\beta$ and $\gamma$ are three pre-defined thresholds.
$\mathcal{F}(\Delta t) = \tau \Delta t$ is a regularization function to control the size of the cache.
Here $\Delta t$ is the duration that an instance is stored in the cache and $\tau$ is the weight.
\begin{equation}
	\left\{
	\begin{aligned}
	&\mathcal{G}_{t,j}^1 & = & ~~ \text{sgn}(p_{t,j} - \alpha)  \\
	&\mathcal{G}_t^2 & = & ~~ \prod_{j=1}^{C}\text{sgn}(\beta - p_{t,j})  \\
	&\mathcal{G}_i^3 & = & ~~ 1 - \prod_{j=1}^{C}\text{sgn}(\gamma - \mathcal{F}(\Delta t) p_{i,j})  \\
	\end{aligned}
	\right.
	,~~
	\text{sgn}(x) = 
	\left\{
	\begin{aligned}
	1&, ~~ \text{if~~} x >= 0, \\
	0&, ~~ otherwise
	\end{aligned}
	\right.
	\label{eq:manual-control}
\end{equation}

\paragraph{\bf A Learnable Controller.}
Designing the gates according to some manually designed rules will highly reduce the generality.
Also, it is hard for us to search for an optimal value of all the hyper parameters.
To make our approach more adaptable, we resort to reinforcement learning (RL) to get a learnable controller. 
RL is characterized by an agent that continuously interacts and learns from the environment through \emph{trial-and-error} games. Its key characteristics include: 1) lack of supervisor, 2) delayed feedback, 3) sequential decisions, and 4) actions affect states, which accord with the peculiarities of our online memory learning setting. 
Specifically, at each time step, we do not know if updating memory can earn long-term benefits; we observe the instances and make the judgments sequentially; and the updating of our memory will influence future judgments. 
RL has the potential to find a better policy to replace naive threshold-based strategy. Here, we take $\mathcal{G}_{t,j}^1$ as an example for analysis.

\noindent\textbf{Problem Formulation.} The game we teach our agent to play is learning a policy $\mu_{\theta}(s)$ to decide whether to update the memory bank. For a new instance $x_t$ with feature representation $f_t$ in the sequential movie, we compare it with $\cM_t$, and repeat this procedure for each cast $j \in \{1, \cdots, C\}$.

\noindent\textbf{State.} State space here is formulated as $\cS_{t} = (\cM_{t}, f_t)$.

\noindent\textbf{Action.} Action space here is a one-dimensional discrete space $\{0,1\}$. 
If action $1$ is taken, we update the memory as Eq.~\ref{eq:mem-update}.

\noindent\textbf{Reward.} Denote the recognition reward at time step $t$ as $r_t$. If the action is matched with the ground truth label, \ie if $x_t$ is indeed the person $j$ and action 1 is taken, or $x_t$ is not $j$ and action 0 is taken, then the recognition reward at the current time step is $r_t = 1$. 
Since the effect of the update can only be reflected in future decisions, 
we define the long-term reward for each action as the cumulative recognition reward in the near future, $R_t = \sum_{m=t}^{t+T} r_{m}$.
We use deep Q-learning network (DQN) to find the improved policy.

The formulation of a learnable $\mathcal{G}_t^2$ is similar to $\mathcal{G}_{t,j}^1$.
Note that we do not employ a learnable $\mathcal{G}_i^3$ here.
The reason is that $\mathcal{G}_i^3$ is dependent on the samples in the UIC, yet the cache size is quite small and unstable, with which we are not able to train an agent.
Through our study, we find that the manual $\mathcal{G}_i^3$ can work well with the other two learned gates. An extensive analysis on the parameters of $\mathcal{G}_i^3$ is provided in the experiment section.

% !TEX root = ../main.tex

\section{Experiments}

\subsection{Experimental Settings}

\noindent\textbf{Data.}
To validate the effectiveness of our approach, we conduct experiments on the state-of-the-art large-scale \emph{Cast Search in Movies} (CSM) dataset~\cite{Huang_2018_ECCV}. Extracted from $192$ movies, CSM consists of a \textit{query} set that contains the portraits of $1,218$ cast (the actors and actresses) and a \textit{gallery} set that contains $127K$ instances (tracklets). The movies in CSM are split into training, validation and testing sets without any overlap of cast. The training set contains $115$ movies with $739$ cast and $79K$ instances, while the testing set holds $58$ movies with $332$ cast and $32K$ instances, and the rest $19$ movies are in the validation set. \\

\noindent\textbf{Evaluation.}
Given a query with the form of a portrait, our method should present a ranking of all the instances in the gallery to suggest the corresponding possibilities that the instances and the query share a same identity. Therefore, we use \textit{mean Average Precision} (mAP) to evaluate the performance. The training, validation and testing are under the setting of \textbf{``per movie"}, 
\ie given a query, a ranking of instances from only the specific movie will be returned, which is in accordance with real-world applications such as ``intelligent fast forwards". Among the 192 movies in CSM, the average size of query and gallery for each movie is 6.4 and 560.5, respectively.

\subsection{Implementation Details}
\noindent\textbf{Multi-modal Feature Representations.} 
For each instance in CSM, we collect face, body and audio features to facilitate multi-modal person search. %Since the method works on an instance-to-instance basis, 
The face and body features are extracted for each frame, and averaged to produce the instance-level descriptors. For body feature, we utilize the IDE descriptor~\cite{ide_2016_ECCV} extracted by a ResNet-50~\cite{resnet_2016_CVPR}, which is pre-trained on ImageNet~\cite{imagenet_2015_IJCV} and finetuned on the training set of CSM. We detect face region~\cite{facedetect_2016_SPL,rao2020unified} and extract face feature with a ResNet-101 trained on MS-Celeb-1M~\cite{msceleb_2016_EI}. NaverNet~\cite{chung2019naver} pre-trained with AVA-ActiveSpeaker dataset~\cite{ava_2019_arxiv} is applied on the instances to align the characters with their speech audio, which distinguishes a character's voice with the others' as well as background noises. With the proper setting of sampling rate and Mel-frequency cepstral coefficients (MFCC)~\cite{logan2000mel} to reduce the noises, ultimately, each speaking instance is assigned with an audio feature. \\

%\vspace{5pt}
\noindent\textbf{Memory Initialization and Update.}
Recall that the multi-modal memory bank is $\cM = \{M_f,M_b,M_a\}$, where $M_{f}$ is initialized with the face features extracted from the IMDb portrait of each actor/actress in the movie, and $M_{b}, M_{a}$ are void. The optimal $\mu$ in Eq.~\ref{eq:mem-update} is set to $0.01$ through grid search.\\

%vspace{5pt}
\noindent\textbf{RL Training.}
The DQN mentioned above is instantiated by a two-layer fully-connected network.
The training epoch is 100 with learning rate 0.001.
Each epoch is run on the whole movie list, with each movie taking 200 Q-learning iterations. The future reward length is set to be 30.
We run the framework on a desktop with a TITAN X GPU.

\begin{table}[!t]
	\begin{center}
		\caption{\label{table:results}
			Person search results on CSM under ``per movie'' setting}
\resizebox{0.9\columnwidth}{!}{%
	\begin{tabular}{l|c|c|c}
	\toprule
	Methods & ~online~ & ~mAP (\%, $\uparrow$)~ & ~~ complexity * ($\downarrow$) \\
	\midrule
	Face matching  & \checkmark & 61.24 & $\mathcal{O}(NC)$ \\ 
	TwoStep~\cite{loy2019wider} (face+body) & & 64.79 & $\mathcal{O}(NC)$\\
	TwoStep~\cite{loy2019wider} (face+body+audio) & & 64.40 & $\mathcal{O}(NC)$ \\
	LP~\cite{zhou2004learning} & & 9.33 & $\mathcal{O}(NC+N^2)$ \\
	PPCC~\cite{Huang_2018_ECCV}   & & 67.99  & $\mathcal{O}(NC+N^2)$ \\
	\midrule
    \textbf{OMS} (DMB w/ manual updating rule) & \checkmark & 63.83  & $\mathcal{O}(NC)$ \\
    \textbf{OMS-R} (DMB w/ RLC) & \checkmark & 64.39 & $\mathcal{O}(NC)$ \\
    \textbf{OMS-RM} (DMB w/ RLC+MFR) & \checkmark & 66.42 & $\mathcal{O}(NC)$ \\
    \textbf{OMS-RMQ} (DMB w/ RLC+MFR+UIC) & \checkmark & \textbf{69.08} & $\mathcal{O}(NC+\hat{k}NC)$ \\
	\bottomrule
	\end{tabular}
}
	\\ \small{* $N$: number of instances; $C$: number of cast; $\hat{k}$: average size of UIC}

\end{center}
\end{table}

\subsection{Quantitative Results}

We compare our method with five baselines: \textbf{1) Face matching (online):} The instances are sequentially compared with the cast portraits by face feature similarity, without memory updating. \textbf{2) TwoStep (face+body):} After comparisons between face features, instances with high recognition confidence are assigned with identity labels, then a round of body feature comparisons is conducted. \textbf{3) TwoStep (face+body+audio): } The second step of comparisons in 2) is based on the combination of body and audio features. \textbf{4) LP:} The identities of labeled nodes are propagated to the unlabeled nodes with conventional linear diffusion~\cite{zhou2004learning} through multi-modal feature links, where a node updates its probability vector by taking a linear combination of vectors from the neighbors~\cite{Huang_2018_ECCV}. In addition to face features, the body and audio features are combined for matching, where the weights are 0.9 and 0.1, respectively. \textbf{5) PPCC}~\cite{Huang_2018_ECCV}\textbf{:} Based on the combination of visual and temporal links, the label propagation scheme only spreads identity information when there is high certainty.

Moreover, four variants of our OMS method are compared to validate the influences of different modules. For \textit{DMB with manual updating rule}, only face features are compared between instances and cast in the memory. When the face similarity exceeds a fixed threshold, the memory is updated with the newly recognized instance. The RL-based controller, multi-modal feature representations and UIC are added sequentially to form the other three variants.

We compare different approaches in three aspects: (1) feasibility of online inference; (2) effectiveness measured by mAP; and (3) computational complexity.

The results are presented in Table~\ref{table:results}, from which we can see that:
1) Almost all the previous works tackle this problem in an offline manner except for the simple face matching baseline,
while OMS can handle the online scenarios.
2) OMS is quite effective, which can even outperform the offline methods significantly.
3) The computational cost of OMS is low. Without UIC, OMS is as efficient as the matching-based methods, \eg face matching.
Note that the cache size $\hat{k}$ is usually smaller than $N$ and $C$ is less than $10$ here. 
Therefore, even for the complete version of OMS, \ie OMS-RMQ, the complexity is still lower than the popular propagation-based methods~\cite{zhou2004learning,Huang_2018_ECCV}.
4) The gradually added components of OMS can continuously raise the performances, which proves the effectiveness of the design for each module.
All these results demonstrate that OMS is an effective and efficient framework for person search.

\subsection{Ablation Studies}
\begin{figure}[!t]
	\centering
	\includegraphics[width=0.98\textwidth]{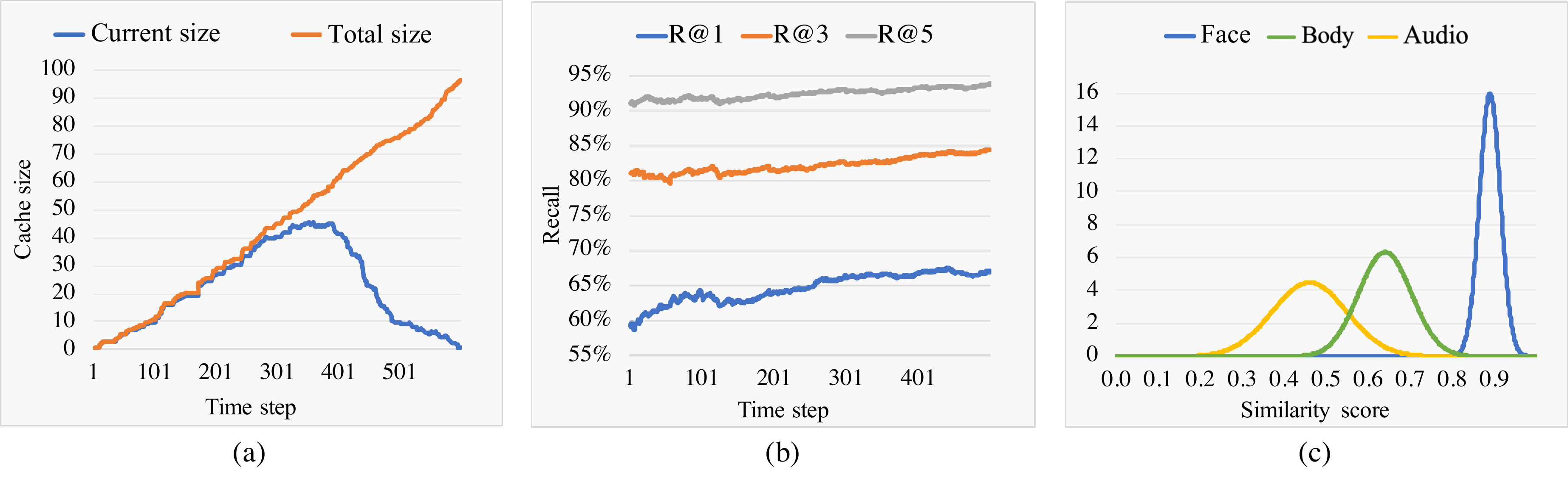}
	\caption{Ablation studies. (a) The variation of cache size along the time. (b) The development of R@k performance along the time. (c) The distribution of similarity scores of the instances that update the DMB}
	\label{fig:analysis}
\end{figure}

\noindent\textbf{What is the behavior of the framework along the time?}
To discover the behavior of our online framework, we study the development of UIC along the time in our OMS-RMQ method.
We select the movies with above 600 instances and record their varied cache sizes at each time step during testing, and average the results among all the movies. 
The result is presented in Figure~\ref{fig:analysis} (a). Each time step represents that a decision is made to an instance, and the first 600 steps are shown. The ``total size'' denotes the cumulative number of uncertain instances that have been put into the cache, while ``current size'' indicates the number of existing instances therein.
It is observed that as time goes, total cache size increases gradually. After processing 600 instances, there are around 100 instances that have ever been put into the UIC.
The current cache size raises at the very beginning. After around 350 time steps, it drops gradually to zero. 
It demonstrates that our DMB becomes better after absorbing informative features to assist recognition, thus more and more uncertain instances get a confident result and are popped out of the cache. 

Additionally, we record the cumulative recall of instance identification results, namely R@k, along the time. R@k means the fraction of instances where the correct identity is listed within the top k results. The performance improves gradually with time, as shown in Figure~\ref{fig:analysis} (b). The R@1 raises from 59\% at the beginning to 67\%, which proves the effectiveness of our online design.\\

\noindent\textbf{What does RL learn?}
Recall that in the manual rule setting, we update the memory if the similarity between the memory and the instance is higher than a given threshold.
With RL, whether to update or not is decided by the trained agent.
To have a deeper understanding of how the RL agent makes the decision and why the RL-trained strategy performs better than manual rules, during testing with OMS-RMQ method, we record the similarity scores on different modalities when an instance is used to update the memory.
After regressing the data points into Gaussian distributions as shown in Figure~\ref{fig:analysis} (c), the mean and standard deviation of similarity scores on face, body and audio are 0.89, 0.025 (face), 0.64, 0.063 (body), and 0.46, 0.089 (audio), respectively. The RL agent implicitly adjusts the thresholds of updating memory. Interestingly, the mean values are almost the same with the thresholds we carefully designed before that achieve the highest performance in the manual rule setting.\\

\begin{figure}[!t]
	\centering
	\includegraphics[width=0.9\linewidth]{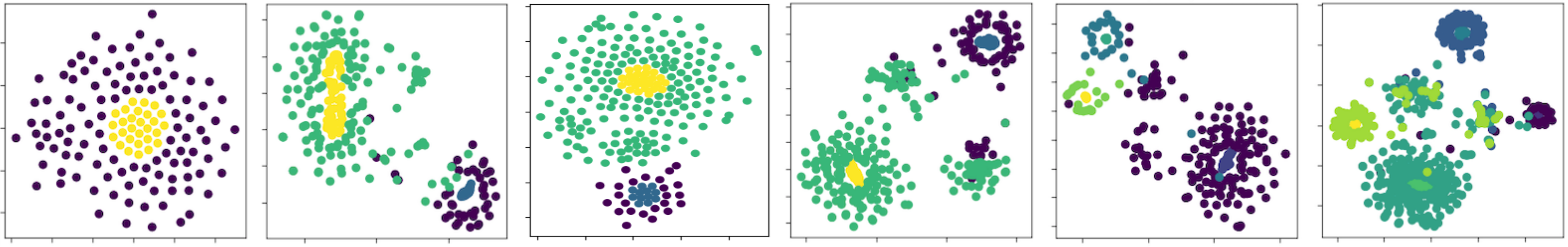}
	\caption{t-SNE plot of instance features and evolution of memory in 6 movies. Each cluster represents a cast. The ``remembered'' features in the memory are plotted by light-colored dots, while the instances of a cast are in dark colors. We notice that all the ``remembered'' features lie at the center of the spread instances. This indicates that the memory absorbs reliable features and well represents the cast's peculiarities}
	\label{fig:tsne}
\end{figure}

\noindent\textbf{What does memory learn?}
To prove the effectiveness of the DMB, we visualize the features of a cast's memory and all his/her ground-truth instances in our OMS-RMQ method using t-SNE~\cite{maaten2008visualizing}. Figure~\ref{fig:tsne} shows cast from 6 movies who have at least 15 memory updates, where each cluster represents a cast. We observe that the updated memory features lie at the center of the instance cluster, which indeed provide typical representations of the cast.
This shows that our DMB can accurately capture the characteristics of all his/her possible lookings.\\

\noindent\textbf{How do different modalities work?}
To study how different modalities contribute to the online multi-modal search, \emph{OMS using DMB with RLC and UIC} is taken as the baseline. The results are shown in Table~\ref{tab:multi}.
The performance improves when we gradually add a new modality information in.
We observe that the introduction of body and audio features brings 3\% and 0.5\% improvement to the baseline, respectively. With all these modalities together, OMS achieves a 4.2\% enhancement in recognition precision, which validates that all the modalities are complementary to each other and are informative to the online search.\\

\begin{table}[!t]
	\caption{Performances of OMS (DMB w/ RLC+UIC) based on different modalities}
	\begin{center}
%		\resizebox{\columnwidth}{!}{
		\begin{tabular}{l|ccc|cc}
			\toprule
			Method           &face &body &audio
		 & mAP (\%, $\uparrow$) \\
			\midrule
			Face matching (\textit{online}, w/o DMB)     &\checkmark&&   & 61.24  \\
			\midrule
			OMS (DMB w/ RLC+UIC)&\checkmark&& 
			& 64.91  \\
 			OMS (DMB w/ RLC+UIC)&\checkmark&\checkmark& 
			& 67.93  \\
  		    OMS (DMB w/ RLC+UIC)&\checkmark&&  \checkmark  
			& 65.39   \\
			OMS (DMB w/ RLC+UIC) &\checkmark&\checkmark&\checkmark 
			& \textbf{69.08} &  \\
			\bottomrule
		\end{tabular}
%				}
	\end{center}
	\label{tab:multi}
\end{table}

\noindent\textbf{What is the effect of different UIC sizes?}
As we mentioned above in $\cG_i^3$, $\mathcal{F}(\Delta t) =\tau \Delta t$ is the regularization function to control the cache size and $\tau$ is the weight.
A larger weight leads to a smaller cache, and vice versa. We select different weights and show the corresponding performances provided by OMS-RMQ in Table~\ref{tab:cache}.
Under each setting, we record the mean cache size of each movie and average the values among all the movies. The average of mean cache size drops from 199 to 16 as the weight raises from 0 to 0.20. The mAP achieves the maximum 69.08\% when the weight is 0.08. 
When the cache is too small, the uncertain instances are not able to benefit from the gradually absorbed knowledge, which causes inferior performance.
Since a character is likely to appear again in the movie before long, a medium-sized cache encourages the uncertain instances to match with a neighboring confidently recognized one. Thus, when the mean cache size is 96, the framework achieves the best result.

\begin{table}[!t]
	\centering
		\caption{Performances of OMS (DMB w/ RLC+MFR+UIC) with different cache sizes}
	\begin{tabular}{l|cccccccc}
		\toprule
		weight $\tau$ & 0      & 0.04   & 0.08  & 0.12   & 0.16   & 0.20    \\\midrule
		mAP (\%, $\uparrow$)    & 66.84 & 68.92 & \textbf{69.08} & 68.13 & 65.67 & 63.69 \\
		Mean cache size    & 199   & 158   & 96   & 63   & 40   & 16  \\
		\bottomrule
	\end{tabular}
	\label{tab:cache}
\end{table}

\subsection{Qualitative Results}

\begin{figure}[!t]
    \centering
    \subfigure[]{
        \includegraphics[width=0.8\textwidth]{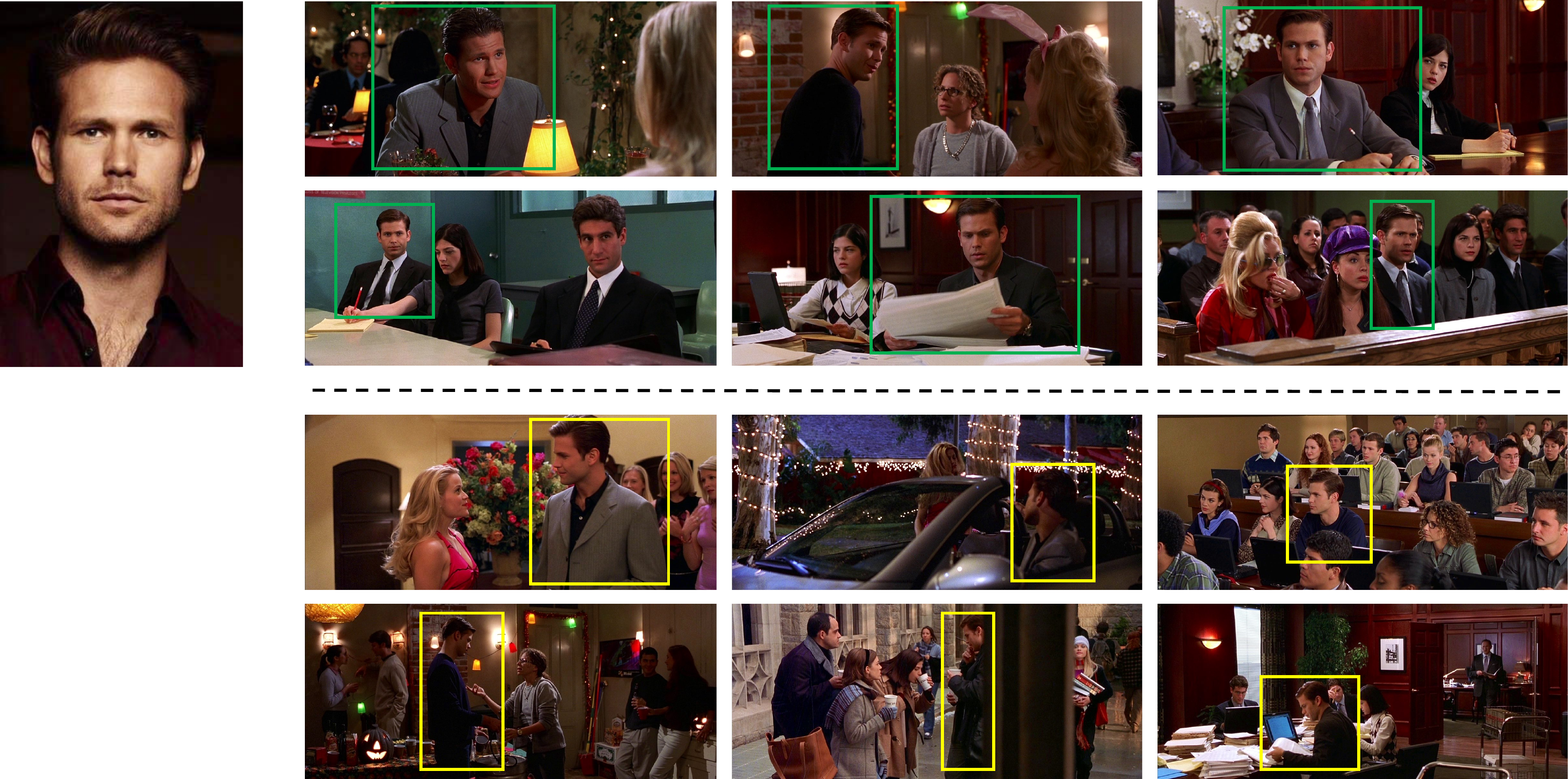}
%        \vspace{-10pt}
    }
    \subfigure[]{
        \includegraphics[width=0.8\textwidth]{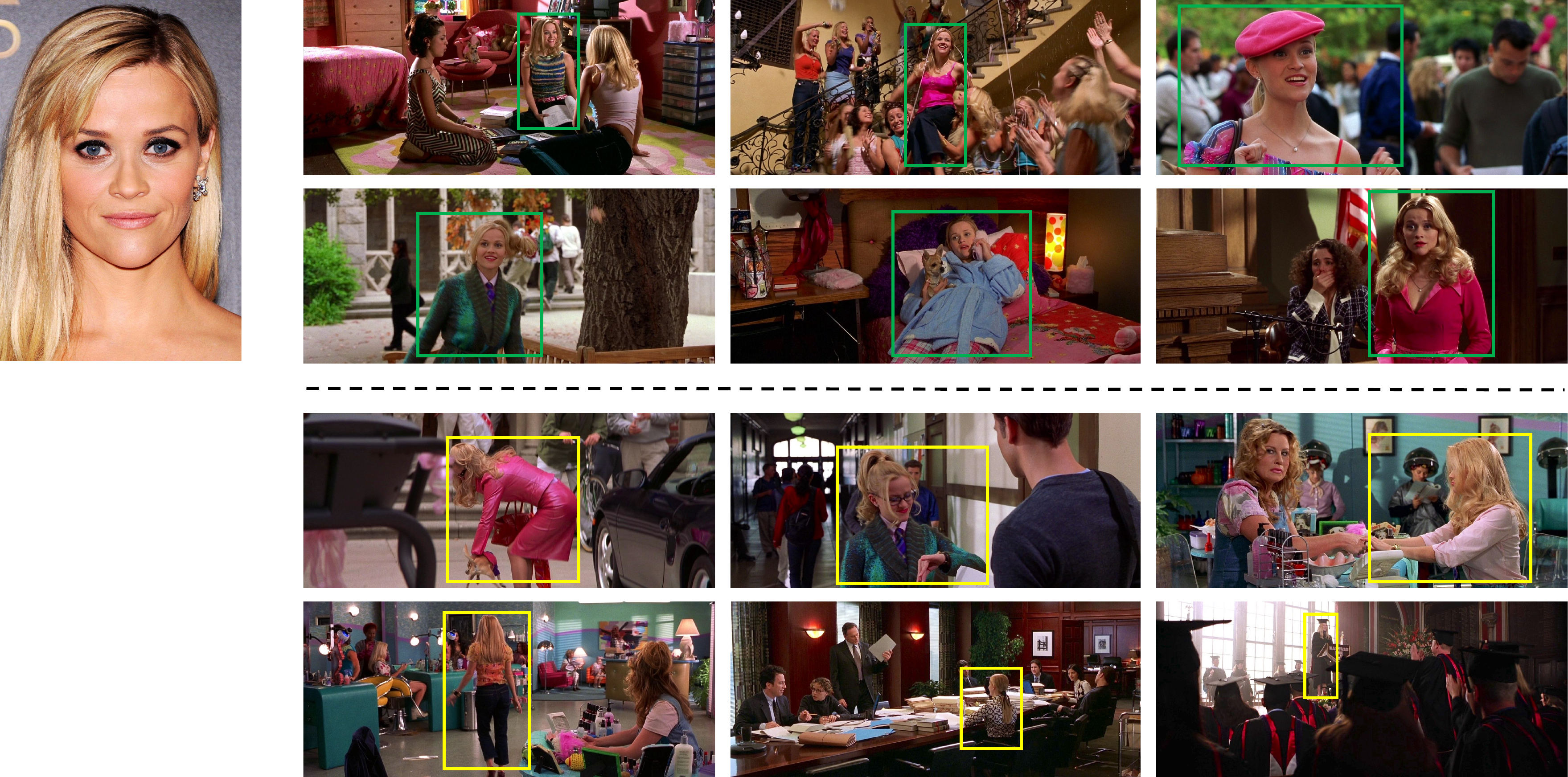}
%        \vspace{-10pt}
    }
%        \vspace{-10pt}
    \caption{The sample instances and their corresponding decision making results given by the trained agent. Samples shown above the dash line with green boxes are well-recognized and used to update memory, while those below the dash line with yellow boxes are temporarily put into the UIC. (a) Movie IMDb ID: tt0250494, cast IMDb ID: nm0205127. (b) Movie IMDb ID: tt0250494, cast IMDb ID: nm0000702}
    \label{fig:memory}
\end{figure}

\noindent\textbf{Which instances contribute to the memory/are sent into the UIC?}
In Figure~\ref{fig:memory}, we present some sample instances and the corresponding actions given by the agent during inference. The samples demonstrate that \textit{person search with one portrait} is extremely challenging due to varied illumination, sizes, expressions, poses and clothing.
During inference, the trained agent successfully selects informative instances which are mostly easier to recognize to update the memory bank, while the instances that contain profile faces, back figures and occlusions are sent into the UIC for later comparisons when more information is acquired.\\

\begin{figure}[!t]
    \centering
    \subfigure[]{
        \includegraphics[width=0.85\textwidth]{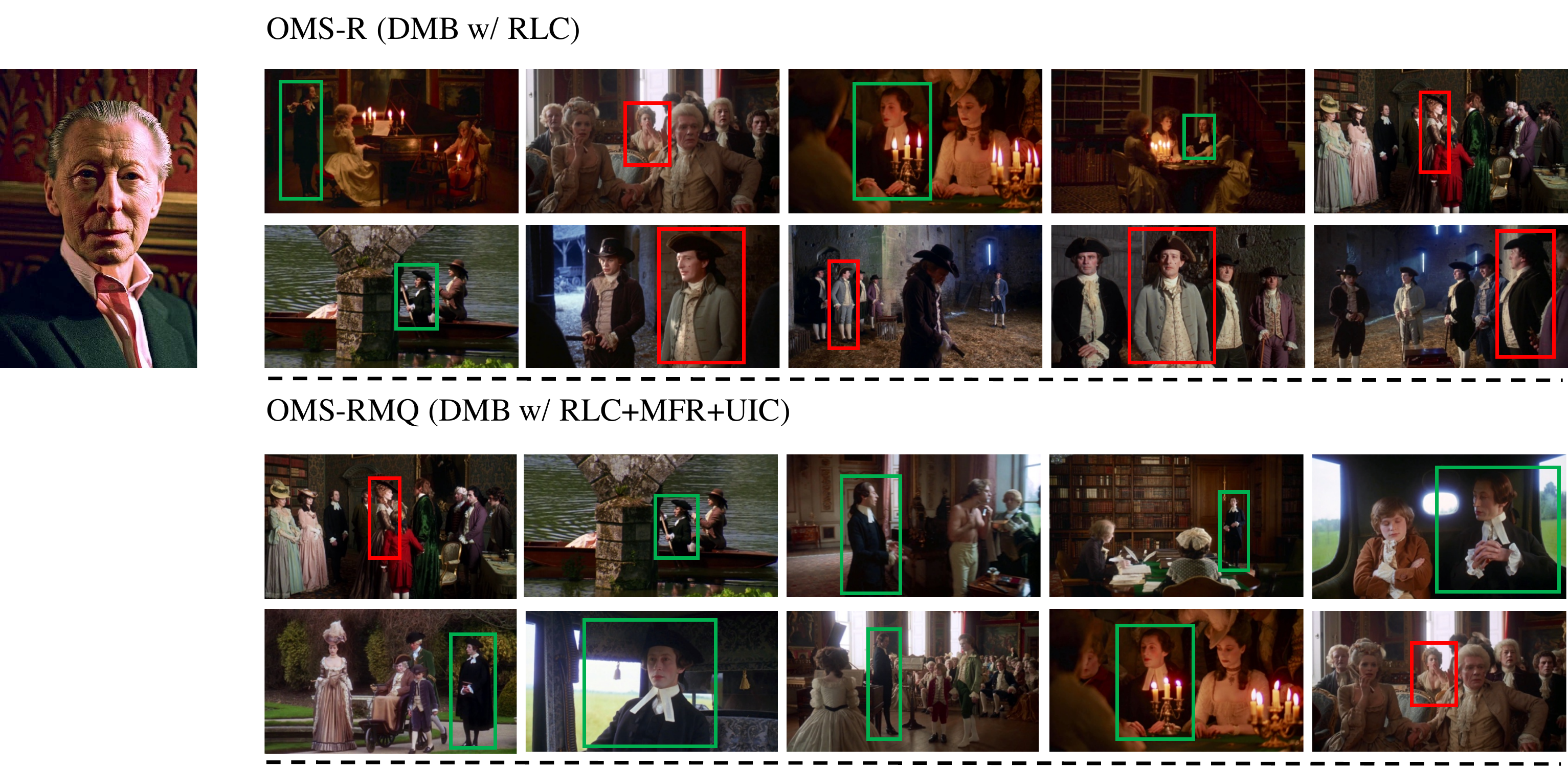}
    }
    \subfigure[]{
        \includegraphics[width=0.85\textwidth]{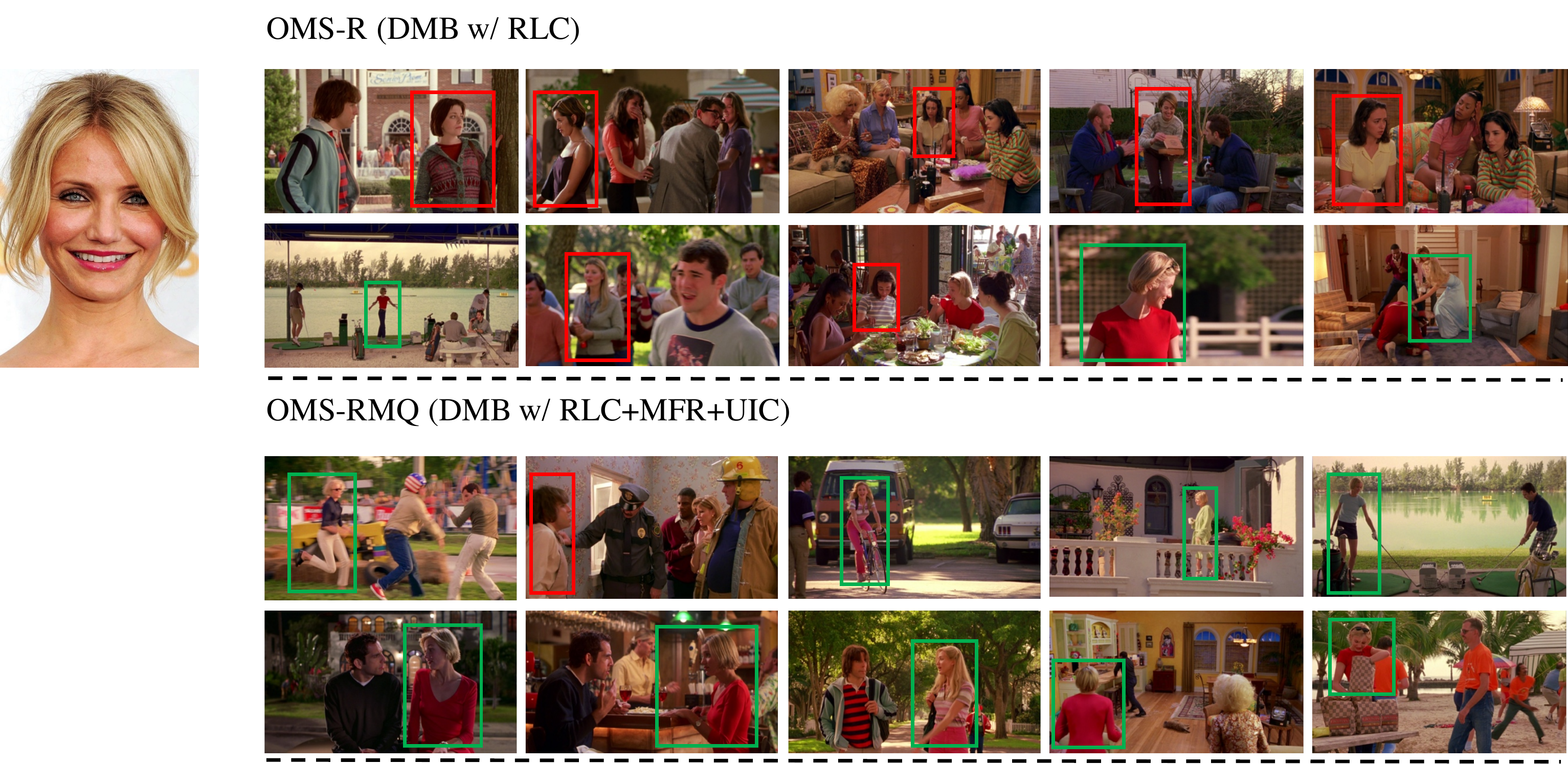}
    }
    \caption{Samples searched by different methods, ranked in descending order according to similarity. The green bounding boxes represent correct recognition, and the red ones are mistakenly identified. (a) The 11th-20th searching results of the actor's portrait. Movie IMDb ID: tt0072684, cast IMDb ID: nm0578527. (b) The 71th-80th searching results of the actress's portrait. Movie IMDb ID: tt0129387, cast IMDb ID: nm0000139}
    \label{fig:compare}
\end{figure}

\noindent\textbf{Method Comparison.}
In the ``per movie'' setting, given a portrait as a query, instances are ranked in descending order according to their similarity to the cast. 
In Figure~\ref{fig:compare}, we show some searching results provided by our OMS-R and OMS-RMQ methods. The green bounding boxes represent correct recognition, while the red ones are mistakenly identified.
It is shown that with the introduction of UIC and multi-modal features, the recognition accuracy is evidently improved, which is in accordance with the quantitative result that the mAP raises from 64.39\% to 69.08\%. 
Even though the rankings of the samples presented are approaching the length of ground-truth instance list, \ie 11-20/22 and 71-80/109, where instances are harder to recognize due to varied poses and face sizes, OMS-RMQ still provides satisfying results.

%\input{articles/application.tex}
% !TEX root = ../main.tex

\section{Conclusion}
\label{sec:conclusion}

In this paper, we systematically study the challenging problem of \textit{person search in videos with one portrait}. To meet the demand of timely inference in real-world video-related applications, we propose an \textit{online multi-modal searching machine}. Inspired by the cognitive process in movie watching experience, we construct a dynamic memory bank to store multi-modal feature representations of the cast, and develop a controller to determine the strategy of memory updating. An uncertain instance cache is also introduced to temporarily keep unrecognized instances for further comparisons. 
Experiments show that our method provides remarkable improvements over online schemes and outperforms offline methods. \\

\noindent\textbf{Acknowledgment}: This work is partially supported by the SenseTime Collaborative Grant on Large-scale Multi-modality Analysis (CUHK Agreement No. TS1610626 \& No. TS1712093), the General Research Fund (GRF) of Hong Kong (No. 14203518 \& No. 14205719), and Innovation and Technology Support Program (ITSP) Tier 2, ITS/431/18F.

\clearpage
\bibliographystyle{splncs04}
\bibliography{main}
\end{document}